\documentclass[12pt]{article}
\usepackage{xr}
\usepackage[legalpaper,  margin=1in]{geometry}

\usepackage{bm}
\usepackage{amsmath,amsfonts,amssymb,amsthm,epsfig, graphicx, hyperref, mathtools, appendix, comment, caption, subcaption, float,setspace, booktabs,natbib,algorithm}
\usepackage[flushleft]{threeparttable}

\usepackage[dvipsnames,table,dvipsnames*, svgnames*, hyperref]{xcolor}

\theoremstyle{plain}
\newtheorem{thm}{Theorem}[section]

\newtheorem{defn}[thm]{Definition}

\theoremstyle{remark}

\newcommand{\beq}{\begin{equation}}
	\newcommand{\eeq}{\end{equation}}
\newcommand{\beas}{\begin{align*}}
	\newcommand{\eeas}{\end{align*}}
\newcommand{\bea}{\begin{align}}
	\newcommand{\eea}{\end{align}}
\newcommand{\bei}{\begin{itemize}}
	\newcommand{\eei}{\end{itemize}}
\newcommand{\ben}{\begin{enumerate}}
	\newcommand{\een}{\end{enumerate}}
\newcommand{\bet}{\begin{theorem}}
	\newcommand{\eet}{\end{theorem}}
\newcommand{\bel}{\begin{lemma}}
	\newcommand{\eel}{\end{lemma}}
\newcommand{\bep}{\begin{proposition}}
	\newcommand{\eep}{\end{proposition}}
\newcommand{\bed}{\begin{definition}}
	\newcommand{\eed}{\end{definition}}
\newcommand{\bec}{\begin{corollary}}
	\newcommand{\eec}{\end{corollary}}
\newcommand{\bex}{\begin{example}}
	\newcommand{\eex}{\end{example}}

\newcommand{\bu}{\bold{u}}

\newcommand{\bw}{\bold{w}}
\newcommand{\bv}{\bold{v}}

\newcommand{\bbf}{\bold{f}}
\newcommand{\bg}{\bold{g}}

\newcommand{\bh}{\bold{h}}

\newcommand{\bU}{\bold{U}}

\newcommand{\bE}{\bold{E}}
\newcommand{\bW}{\bold{W}}

\newcommand{\bV}{\bold{V}}

\newcommand{\bH}{\bold{H}}

\newcommand{\bY}{\bold{Y}}

\newcommand{\bX}{\bold{X}}

\newcommand{\bS}{\bold{S}}

\newcommand{\bF}{\bold{F}}

\newcommand{\bTheta}{\bold{\Theta}}

\newcommand{\R}{\mathbb{R}}
\newcommand{\E}{\mathbb{E}}

\newcommand{\vertiii}[1]{{\left\vert\kern-0.25ex\left\vert\kern-0.25ex\left\vert #1 
		\right\vert\kern-0.25ex\right\vert\kern-0.25ex\right\vert}}

\newcommand{\fb}{\mathbf{f}}

\usepackage{amsmath}
\usepackage{amsfonts}
\usepackage{amssymb}
\usepackage{algpseudocode}

\title{U-aggregation: Unsupervised     Aggregation of Multiple Learning Algorithms}
\author{Rui Duan\\
    Department of Biostatistics\\
    Harvard T.H. Chan School of Public Health \\ 
    Boston, MA 02115, USA
    }

\date{}

\begin{document}

\maketitle
\doublespacing 

\begin{abstract}
      Across various domains, the growing advocacy for open science and open-source machine learning has made an increasing number of models publicly available. These models allow practitioners to integrate them into their own contexts, reducing the need for extensive data labeling, training, and calibration. However, selecting the best model for a specific target population remains challenging due to issues like limited transferability, data heterogeneity, and the difficulty of obtaining true labels or outcomes in real-world settings. In this paper, we propose an unsupervised model aggregation method, U-aggregation, designed to integrate multiple pre-trained models for enhanced and robust performance in new populations. Unlike existing supervised model aggregation or super learner approaches, U-aggregation assumes no observed labels or outcomes in the target population. Our method addresses limitations in existing unsupervised model aggregation techniques by accommodating more realistic settings, including heteroskedasticity at both the model and individual levels, and the presence of adversarial models. Drawing on insights from random matrix theory, U-aggregation incorporates a variance stabilization step and an iterative sparse signal recovery process. These steps improve the estimation of individuals' true underlying risks in the target population and evaluate the relative performance of candidate models. We provide a  theoretical investigation and systematic numerical experiments to elucidate the properties of U-aggregation. We demonstrate its potential real-world application by using U-aggregation to enhance genetic risk prediction of complex traits, leveraging publicly available models from the PGS Catalog.\\
      
\end{abstract}
\newpage
\section{Introduction}

With the growing emphasis on open science and open-source machine learning, alongside the continuous accumulation of new data across diverse fields, pre-trained models are increasingly available for public use \citep{sonnenburg2007need,mckiernan2016open}. This shift enables researchers and practitioners to leverage state-of-the-art models without the need for extensive computational resources or large-scale datasets for training. For example, in the field of polygenic risk prediction—where models are designed to predict individuals' phenotypes based on their genotypes—the PGS Catalog was launched as an open-access database, allowing researchers to directly download and apply existing Polygenic Risk Score (PRS) models in their studies \citep{choi2020tutorial,lambert2021polygenic}. In just a year or two, over 5,000 models covering more than 650 traits have become publicly available, with the numbers growing rapidly. For common traits such as height, body mass index, and breast cancer, there are over 100 pre-trained models available, developed by different research groups using a variety of statistical and machine learning methods across diverse datasets. These publicly available models foster transparency, reproducibility, and collaboration in machine learning, offering new opportunities for research in fields that benefit from model-sharing and the integration of diverse data sources \citep{lambert2024polygenic}.

When applying pre-trained models to a new population, model accuracy often decreases due to distributional shifts between the source and target populations \citep{cai2023diagnosing}. This creates challenges for practitioners, who may face uncertainties regarding the model's reliability in their specific population of interest. Common approaches to address this issue include refitting or calibrating existing models through techniques such as transfer learning, fine-tuning, or model recalibration, which adjust model parameters to better match the target population's characteristics \citep{weiss2016survey,huang2020tutorial}. These methods, however, typically require either the access to the data used to train the source model, or a validation dataset from the target population with outcome measures or labels \citep{guo2017calibration,xiong2023distributionally,gu2022robust,li2022transfer,yuan2024optimal,hector2024turning}. Furthermore, they are generally designed to calibrate individual models, making them less practical when a large number of pre-trained models are available for use.

As pre-trained models become more widely available, effective methods for combining them have gained importance, enabling the strengths of individual models to be harnessed for improved overall performance.  For example, a series of work has considered supervised aggregation of classifiers or models, which produces certain types of optimal models, e.g., the best among all linear combinations of pre-trained models \citep{lecue2014optimal,tsybakov2004optimal}. Related approaches include classical ensemble learning techniques such as boosting \citep{schapire2003boosting}, bagging \citep{breiman1996bagging}, and the super learner methods \citep{van2007super,dong2020survey}, which typically involve training multiple models and aggregating them  from the same dataset with observed outcomes or labels. With pre-trained models now readily available, these methods can be adjusted to focus solely on model aggregation, eliminating the need to train individual learners from scratch.  

A key limitation of the model aggregation methods aforementioned is that, in practical implementation, supervised training may not be feasible when labels or outcome measures are unavailable. For example, evaluating genetic risk scores for a particular disease in a relatively young population can be challenging, as the majority may not yet have experienced disease onset. Additionally, accurately labeling disease status from observational data, such as electronic health records, can be costly, often requiring specialized clinical expertise \citep{lobach2007research}. Furthermore,  the aggregated model’s performance is tied to the specific training data used, which can limit its transferability and may result in sub-optimal performance when applied to new datasets. Targeting the model transferability,  methods such as maximin learning, environment-invariant learning, distributionaly robust learning, were proposed in the recent years for training a model across data from heterogeneous  environments, with the hope that it can be more transferable to unseen new environments \citep{maximin2015,rahimian2019distributionally,buhlmann2020invariance,guo2024statistical}. However, these methods require data from the source populations, which are less available in practice compared to the fitted model itself. 
As a consequence, potential strategies for unsupervised model aggregation which directly aggregate predicted values from pre-trained models can be highly practically useful.

There are a few work focusing on unsupervised  model aggregation. Specifically, \cite{parisi2014ranking} proposed a  meta-classifier that integrates a set of binary classifiers, assuming conditional independence between classifiers.\cite{ahsen2019unsupervised} extended \cite{parisi2014ranking} to consider integrating models that output continuous scores instead of binary classification. \cite{ma2023spectral} considered integrating multiple dimension reduction algorithms, assuming output of each algorithm can be written as signal plus homoskedastic Gaussian noises. \cite{coombes2020principal} proposed to use principal component analysis (PCA) to improve tuning parameter choices in PRS models. All the aforementioned methods involve applying spectral decomposition to a concordance matrix constructed from output of all classifiers or algorithms, and obtain a consensus output through a weighted average of all models where the weights are determined by the eigenvectors. Additionally, a series of studies have focused on unsupervised rank aggregation, where multiple ranking functions or rankers provide ranked or partially ranked lists of subjects, with the goal of learning a consensus ranking across all functions \citep{volkovs2014new,klementiev2008unsupervised}. This is closely related to the field of crowdsourcing, which finds broad applications in tasks such as image labeling, text categorization, and sentiment analysis. In these contexts, large groups of individuals, often participating via online platforms, contribute annotations that are used to train and enhance machine learning models \citep{ghezzi2018crowdsourcing,vaughan2018making,xu2024crowdsourcing}.
However, when dealing with algorithms that produce continuous outputs, traditional rank aggregation methods may not be directly applicable or may not be able to fully leverage the magnitude of these outputs.

Unlike existing unsupervised model aggregation approaches, we address some more realistic and challenging scenarios in aggregating multiple models. Specifically, we account for the possibility that candidate models exhibit varying levels of performance on the target dataset, leading to heteroskedastic noise in the model-specific predictions. Additionally, we reason that models may perform differently in the prediction accuracy across subjects, as the risk for certain subjects may be more challenging to estimate than  others. Lastly, we acknowledge that practitioners might include entirely non-informative models, either due to training issues or reporting errors when models are made publicly available. Given these complexities, previous methods often yield inconsistent results; see Section \ref{sec.simu} for numerical results. In this study, we propose a novel spectral approach, named \textit{U-aggregation}, for unsupervised model aggregation. The proposed method consists of two main steps. The first step involves a data-normalization procedure that stabilizes the potentially heteroskedastic noises, leveraging recent advances from random matrix theory \citep{erdHos2019random,ajanki2019quadratic,landa2022biwhitening,landa2023dyson}. The second step combines a modified power iteration algorithm,  also known as the approximate message passing \citep{donoho2009message,bayati2011dynamics,montanari2021estimation}, and a final renormalization procedure, to obtain estimators of a sparse loading coefficient vector and the true risk values for all subjects. Our theoretical analysis justifies both steps of the proposed methods, establishes their consistency, and reveals the asymptotic behaviors of the aggregated risk score estimators in the high-dimensional regime where both the sample size and the model size increase. Simulation studies show that our method outperforms existing approaches, particularly in heteroskedastic noise settings. We implement \textit{U-aggregation} for integration of pretrained PRS models from the PGS catalog to predict complex human traits within the All of Us (AoU) cohort.


\section{Method}
\subsection{Notation and problem set-up}

We use bold lowercase letters to represent vectors and bold uppercase letters to represent matrices. For a vector $\mathbf{a} \in \mathbb{R}^d$, $a_i$ denotes its $i$-th entry, and $\text{diag}(\mathbf{a})$ represents the $d \times d$ diagonal matrix with diagonal entries equal to $\mathbf{a}$. For a matrix $\mathbf{Y} \in \mathbb{R}^{d \times n}$, $\mathbf{Y}_i$ denotes its $i$-th column or row, as specified by the context.

We consider applying a total of $d$ pre-trained models to a set of $n$ subjects, from which we obtain their predicted values.  Inspired by evidence from real-worlds applications, we propose the following scale-free sparse signal-plus-noise model to capture the structure of the predicted values. We assume $\bY_i\in\R^{n}$, the predicted values for $n$ subjects associated to the $i$th model, satisfy
\beq
\bY_i=c_i(\bv u_i+\sigma_i\bF\bw_i),\qquad i=1,2,...,d,
\label{datamodel}
\eeq
where $c_i$ is a model-specific global scaling factor, $\bv \in \mathbb{R}^n$ is the signal vector capturing the true underlying risk of the $n$ subject, and $u_i$ is a binary variable indicating whether  $\bY_i$ contains information about the signal vector $\bv$, with $s = \sum_{i=1}^d 1\{u_i \neq 0\} \leq d$ indicating the sparsity of the ${u_i}$ values, allowing the presence of a portion of non-informative models. In addition, $\sigma_i\bF\bw_i$ is the noise vector, where $\bF = \text{diag}(\fb^{1/2})$ for some $\fb=(f_1,f_2,...,f_n) \in \mathbb{R}^n$ characterizes the sample-specific noise levels,  $\sigma_i$ denotes the model-specific noise level, and $\bw_i \sim N(0, \frac{1}{n} {\bf I})$.
For identifiability, we normalize $u_i$ such that  $u_i \in \{0, \sqrt{\gamma/s}\}$ for some constant $\gamma>0$, and $\bu = (u_1, \dots, u_d)^\top$ therefore satisfies $\|\bu\|_2 = \gamma$.

In our context, the predicted values from each pre-trained model may vary significantly in scale, as captured by $c_i$, and in performance, as reflected by $u_i$ and $\sigma_i$.  Larger values of $\sigma_i$ correspond to reduced accuracy, while the indicator $u_i$ allows for the inclusion of algorithms that may fail to capture any aspect of the ground truth. This model setup is grounded in practical observations. In the context of aggregating pre-trained PGS models, we observe strong evidence that the predicted values from these models exhibit scaling difference, a rank-one structure as well as various types of heteroskedasticity, as elaborated in Section \ref{sec.model}.



Our goal is to obtain a consensus estimator for $\bv$ that effectively combines the predictions $\{\bY_i\}_{1 \le i \le d}$ from the $d$ pre-trained models, and also to gauge the relative performance of each model $i$, despite the unobserved true underlying risk $\bv$. Achieving these objectives is challenging under the assumed data model in (\ref{datamodel}). First, the relevant models with $u_i = 1$ exhibit heteroskedastic noise at both the sample and model levels, even though they capture the same underlying signal vector $\bv$. Additionally, the inclusion of irrelevant models (where $u_i = 0$) introduces further complexity, making it difficult to distinguish the true signal from noise in the presence of such models. 

To gain a deeper understanding of the problem and to better characterize our objectives, we begin by presenting some insightful analyses. Due to the  scaling differences as reflected in $c_i$, we define $\bar\bY_i=\frac{\bY_i}{\|\bY_i\|_2}$ to be the normalized predicted values of model $i$, and define $\bar\bY\in\R^{d\times n}$ as the matrix containing all $\bar\bY_i$'s.  We observe  that
\beq\label{norm.eq}
\bar\bY=\bar\bu\bv^\top+\bH\bW\bF,
\eeq
where 
\[
\bar\bu=\bigg(\frac{u_1}{\|\bv u_1+\sigma_1\bw_1 \|_2}, ..., \frac{u_d}{\|\bv u_d+\sigma_d\bw_d \|_2}\bigg)^\top,
\]
\[
\bH = \text{diag}\bigg(\frac{\sigma_1}{\|\bv u_1+\sigma_1\bw_1 \|_2},  ..., \frac{\sigma_d}{\|\bv u_d+\sigma_d\bw_d \|_2} \bigg),
\]
and $\bW$ has i.i.d. entries from $N(0,1/n)$. Let $\bE=(E_{ij})=\bH\bW\bF$. We further observe that if we define $\lambda$ as the limit of $\|\bv\|_2$, or $\lim_{n\to\infty}\|\bv\|_2=\lambda$,
it follows that
\[
\|\bv u_i+\sigma_i\bw_i \|_2^2=\|\bv\|_2^2u_i^2+\sigma_i^2\|\bw_i\|_2^2+2u_i\sigma_i\bv^\top\bw_i\to_P \lambda^2u_i^2+\sigma_i^2,
\]
by Law of Large Number. In other words, as $n$ grows, the $i$-th entry of $\bar \bu$ is approximately 
\beq \label{ubar}
\bar u_i \approx \frac{u_i}{\sqrt{\lambda^2u_i^2+\sigma_i^2}}.
\eeq
We see that $\bar \bu$ actually quantifies the performance of each model: for models with $u_i\ne0$, larger  $\sigma_i^2$ would lead to smaller $\bar u_i$; when $u_i = 0$, we have $\bar u_i = 0$. Thus, a good estimator of $\bar \bu$ can be used to evaluate the performance of each model, and  the true risk $\bv$ is contained in the rank-one signal component $ \bar\bu\bv^\top$, which can be recovered simultaneously with $\bar \bu$.

When $\bH$ and $\bF$ are both identity matrices, the problem simplifies to a low-rank (or rank-one) matrix estimation problem under additive homeoskedastic noise. This justifies the use of spectral methods such as PCA or SVD, as employed in prior unsupervised aggregation studies \citep{coombes2020principal,ma2023spectral}. However, the heteroskedastic noise matrix $\bE$ now has rather a complex structure due to normalization, as well as the sample-level and model-level heterogeneity. As shown in prior work \citep{zhang2022heteroskedastic} and further demonstrated in our simulation study, without properly addressing the heteroskedasticity,  applying existing spectrum decomposition methods can result in low accuracy.






\subsection{Variance stabilization using Dyson Equalizer}

There are previous work aiming at estimating the low-rank signal of a matrix in the presence of heteroskedastic noise \citep{zhang2022heteroskedastic,landa2022biwhitening,hong2021heppcat}. However, these methods generally assume independent, mean-zero noise, which does not directly apply to our problem. Furthermore, some approaches are tailored to specific data types or leverage properties of particular data-generating models that do not hold in our case.

Under our data model (\ref{datamodel}), we see that if we define 
\beq \label{h0}
\bh_0=\bigg(\frac{\sigma_1^2}{\lambda^2u_1^2+\sigma_1^2},  ..., \frac{\sigma_d^2}{\lambda^2u_d^2+\sigma_d^2} \bigg)
\eeq
and define 
$\bS=(S_{ij})\in\R^{d\times n}$ such that 
\beq\label{eq:S_rank1}
\bS = \frac{1}{n}\bh_0{\fb}^\top.
\eeq
Then we can see that 
\beq\nonumber
|S_{ij}-\text{Var}(E_{ij})|\to_P 0.
\eeq
This implies that, the element-wise variance matrix $\bS$ of the noise matrix $\bE$ would asymptotically have a rank-one structure, despite its heteroskedasticity.  

This property motivates the stabilization of variance, which we can effectively leverage. With a slight abuse of notation, if a general \( d \times n \) matrix \( \bY = \bX + \bE \), satisfies that \( \bX \) is deterministic low-rank and \( \bE \) is element-wise random independent mean-zero noise with  element-wise variance matrix \( \bS\in\R^{d\times n} \) satisfying
$
\bS = \mathbf{a}\mathbf{b}^\top
$
with some positive vectors $\mathbf{a}\in \R^d$ and $\mathbf{b}\in \R^n$, we can normalize (or bi-whiten) the rows and columns of \(\mathbf{Y}\) by defining 
\[
\tilde\bY =D_{\mathbf{a}}^{-1/2} \bY D_{\mathbf{b}}^{-1/2} = \tilde{\bX} + \tilde{\bE},
\]
where \(D_{\mathbf{a}} = \text{diag}(\mathbf{a})\), \(D_{\mathbf{b}} = \text{diag}(\mathbf{b})\) and
$\tilde{\bX} = D_{\mathbf{a}}^{-1/2} \bX D_{\mathbf{b}}^{-1/2}$ is the normalized signal while $\tilde{\bE} = D_{\mathbf{a}}^{-1/2} \bE D_{\mathbf{b}}^{-1/2}$ is the normalized noise. The resulting matrix noise $\tilde{\mathbf{E}}$ will be homoskedastic, which  allows traditional spectrum decomposition to achieve greater accuracy in signal recovery. Estimating the rank-one structure of the variance of the noise, i.e., $\mathbf{a}$ and $\mathbf{b}$ becomes the key for variance stabilization.


{The Dyson Equation is an important tool for connecting the entries of a random matrix and the variance matrix, which has been used in prior work for variance stabilization \citep{landa2023dyson}.} We borrow similar ideas to design our variance stabilization procedure described in Algorithm 1 and we briefly introduce the high-level idea below. 

Define the symmetrized matrices
\[
\mathcal{Y} = \begin{bmatrix}
   \mathbf{0}_{d\times d} & \bY \\
   \bY^\top & \mathbf{0}_{n\times n}
\end{bmatrix}, \text{ and }\quad \mathcal{S} = \begin{bmatrix}
   \mathbf{0}_{d\times d} & \bS \\
   \bS^\top & \mathbf{0}_{n\times n}
\end{bmatrix}.
\]
The resolvent of \( \mathcal{Y} \) is defined as
\beq\label{res}
\mathcal{R}(z) = (\mathcal{Y} - zI)^{-1},
\eeq
where \( I \) is the identity matrix and \( z \in \mathbb{C}^+ \) is a point in the complex upper half plane \( \mathbb{C}^+ \). Existing works \citep{erdHos2019random,ajanki2019quadratic,landa2023dyson} have established that the main diagonal of \( \mathcal{R}(z) \) would concentrate around the solution \( \mathbf{q} \in \mathbb{C}^{d+n} \) to the following deterministic vector-valued Dyson equation
\beq \label{g1g2}
z + \mathcal{S}\mathbf{q} = -\frac{1}{\mathbf{q}},
\eeq
for any \( z \in \mathbb{C}^+ \). 
Moreover,  choosing $z = i\eta$ for some positive real number $\eta>0$, equation (\ref{g1g2}) reduces to
\[
\eta + \bS \mathbf{g}^{(2)} = \frac{1}{\mathbf{g}^{(1)}}, \quad
\eta + \bS^T \mathbf{g}^{(1)} = \frac{1}{\mathbf{g}^{(2)}},
\]
with ${\mathbf{g}^{(1)}}\in \R^d$ and ${\mathbf{g}^{(2)}}\in \R^n$  and
$\mathbf{g} = ({\mathbf{g}^{(1)}}^\top,{\mathbf{g}^{(2)}}^\top)^\top$ being the  imaginary part  of $\mathbf{q}$ when $z=i\eta$ \citep{erdHos2019random,ajanki2019quadratic,landa2023dyson}. This essentially describes a connection between diagonal values of the resolvent of $\mathcal{Y}$ and the variance matrix  $\bS$ of the noise, that is,
\beq\label{approx}
\text{Im}(\text{diag}(\mathcal{R}(i\eta))) \approx \bg,\qquad \forall \eta>0,
\eeq
which allow us to first estimate $\bg$ using the data matrix $\bY$ and then  gauge the magnitude of $\bS$. In particular, when it further holds that $\bS = \mathbf{a}\mathbf{b}^\top$, the relationship between $\mathbf{a}$, $\mathbf{b}$, and  $\mathbf{g}^{(1)}$,  $\mathbf{g}^{(2)}$ can be established as 
\begin{equation}\label{rank_one_dyson}
\mathbf{a} = \frac{\kappa}{\sqrt{d - \eta \| \mathbf{g}^{(1)} \|_1}} \left( \frac{1}{\mathbf{g}^{(1)}} - \eta \right), \quad
\mathbf{b} = \frac{\kappa^{-1}}{\sqrt{n - \eta \| \mathbf{g}^{(2)} \|_1}} \left( \frac{1}{\mathbf{g}^{(2)}} - \eta \right), \quad
\kappa = \sqrt{\frac{\mathbf{a}^T \mathbf{g}^{(1)}}{\mathbf{b}^T \mathbf{g}^{(2)}}},
\end{equation}
which hold for any constant $\kappa>0$.
In other words,  given that $\mathbf{g}^{(1)}$ and $\mathbf{g}^{(2)}$ can be estimated from the data (via the resolvent $\mathcal{R}(z)$),
$\mathbf{a}$ and $\mathbf{b}$ are now identifiable and can be estimated based on (\ref{rank_one_dyson}), up to a scaling factor $\kappa$. Note that the above discussion holds for any $\eta>0$.

Notably, in our problem, the normalized data matrix $\bar \bY$ defined in equation (\ref{norm.eq}) is not strictly a deterministic low-rank signal matrix plus independent mean-zero noises. However, as will be seen in our theoretical analysis (Section \ref{sec.theory}), it shares similar properties as $n$ grows. Leveraging this idea, we propose to stabilize the variance via Algorithm \ref{alg:variance_stabilization}.

\begin{algorithm}[H]
\caption{Variance Stabilization}
\label{alg:variance_stabilization}

\textbf{Input:} {$\bar{\bY}\in\R^{d\times n}$. Here we assume $d\le n$ without loss of generality.}


\begin{enumerate}
    \item Compute the singular value decomposition  of $\bar{\bY}$:
    \[
    \bar{\bY} = \bU \mathbf{\Theta} \bV^\top
    \]
    where $\bU \in \mathbb{R}^{d \times d}$, $\bV \in \mathbb{R}^{n \times d}$ are the left and right singular vectors, and $\bTheta=\text{diag}(\theta_1, \theta_2,...,\theta_d)$ are the singular values. 
    \item Let $\bar\theta$  be the median of the singular values $\{\theta_i\}_{1 \leq i \leq d}$ of $\bar{\bY}$, and obtain estimators 
\[
    \hat{\bg}^{(1)} = \left(\sum_{k=1}^d \frac{\bar\theta}{\theta_k^2 + \bar\theta^2} U_{ik}^2 \right)_{1 \leq i \leq d}, \quad
    \hat{\bg}^{(2)} = \left(\frac{1}{\bar\theta} + \sum_{k=1}^d \left(\frac{\bar\theta}{\theta_k^2 + \bar\theta^2} - \frac{1}{\bar\theta} \right) V_{ik}^2\right)_{1 \leq i \leq n},
    \]
    where $U_{ik}$ and $V_{ik}$ are $(i,k)$-th entry of $\bU$ and $\bV$, respectively. 

\item Obtain:
    \beq \label{h.hat}
    \hat{\bh} = (\hat{h}_1, \ldots, \hat{h}_d) = \frac{1}{\sqrt{d - \bar\theta \|\hat{\bg}^{(1)}\|_1}} \left(\frac{1}{\hat{\bg}^{(1)}} - \bar\theta\right),
    \eeq
    \beq\label{f.hat}
    \hat{\fb} = (\hat{f}_1, \ldots, \hat{f}_n) = \frac{1}{\sqrt{n - \bar\theta \|\hat{\bg}^{(2)}\|_1}} \left(\frac{1}{\hat{\bg}^{(2)}} - \bar\theta\right).
    \eeq
    
    \item Obtain:
    \[
    \tilde{\bY} = \hat{\bH}^{-1} \bar{\bY} \hat{\bF}^{-1}\in\R^{d\times n},
    \]
    where
    \[
    \hat{\bH} = \text{diag}(n^{1/4} \hat{\bh}^{1/2}), \quad
    \hat{\bF} = \text{diag}(n^{1/4} \hat{\fb}^{1/2}).
    \]
\end{enumerate}

\textbf{Output:} $\tilde{\bY}$. \\
\hrulefill
\end{algorithm}




{In Algorithm \ref{alg:variance_stabilization}, the key step is to utilize the approximation (\ref{approx}) to obtain estimators $\hat{\bg}^{(1)}$ and $\hat{\bg}^{(2)}$  based on the diagonals of the matrix resolvent $\mathcal{R}(z)$ defined in (\ref{res}) with $z=i\bar\theta$, and then leverage Equation (\ref{rank_one_dyson}) to estimate the variance stablization factors $\hat\bh$ and $\hat\fb$.} As will be seen in Section \ref{sec.theory}, the normalizing vectors $\hat\bh$ and $\hat\fb$ obtained from Step 2 are actually consistent estimators, up to possible rescaling, of the rank-one consisting components $\bh_0$ and $\fb$ of the asymptotic variance matrix $\bS$ of $\bE$. As such,
after a bi-whitening procedure, we have 
\beq \label{Ytilde}
\tilde \bY=\hat\bH^{-1}\bar\bY\hat\bF^{-1}=\hat\bH^{-1}\bar\bu\bv^\top\hat\bF^{-1}+\hat\bH^{-1}\bH\bW\bF\hat\bF^{-1},
\eeq
which is the sum of normalized a rank-one signal structure and a normalized noise matrix with relatively homogeneous variances. We denote $\tilde \bY = \tilde \bu_n\tilde \bv_n^\top+\tilde \bE$, where $\tilde \bu_n = \text{diag}(n^{-1/4}\bh^{-1/2})\bar \bu$, 
$\tilde \bv_n = \text{diag}(n^{-1/4} \fb^{-1/2})\bv$ and $\tilde \bE=\hat\bH^{-1}\bH\bW\bF\hat\bF^{-1}$ is the normalized noises.

\subsection{Signal recovery using approximate message passing}

An important feature of our problem is the binary nature of the model-informativeness indicator $\{u_i\}$, which posits that  $\bu$ is a sparse vector. To achieve enhanced estimation of $\bu$ and therefore  enhanced estimation of $\bv$, we propose to use an Approximate Message Passing (AMP) algorithm, that exploits the sparse structure of $\bu$ and recursively update and enhance the estimates of $\bu$ and $\bv$ through a debiased power iteration.

AMP are iterative algorithms motivated by ideas in information theory \citep{richardson2008modern} and statistical physics \citep{mezard1987spin}. By alternating a matrix multiplication and a nonlinear operation on the same vectors, it can achieve efficient parameter estimation within polynomial time. Importantly, the general form of the algorithms allow for sharp characterization of their behavior in the high-dimensional limit, known as ‘state evolution’ \citep{donoho2009message,bayati2011dynamics}, which offers rich insights such as their statistical  properties in parameter estimation (see Section \ref{sec.amp.theory}). In our context of singular vector estimation with  a sparsity constraint, the AMP algorithm resembles the classical projected power iteration \citep{onaran2017projected,chen2018projected}. We present our algorithm as following.


\begin{algorithm}[H]
\caption{Iterative Estimation of $\bar{\bu}$ and $\bv$}
\label{alg:iterative_estimation}
\textbf{Input:}{$\tilde{\bY}$, $\hat\bF$ and $\hat\bH$ from Algorithm 1, a sequence of threshold parameters $\{\tau_t\}_{t \leq L}$.}

\textbf{Step 1: Initialization.}
\begin{itemize}
    \item Set $\tilde{\bv}^0 = \sqrt{n}\hat{\bv}_n$, where $\hat{\bv}_n$ is the first right singular vector of $\tilde{\bY}$.
    \item Initialize $\tilde{\bu}^t = 0$ for $t = -1$.
\end{itemize}

\textbf{Step 2: Approximate Message Passing Iterations.}
\begin{itemize}
    \item For $t = 0, 1, \ldots, L$:
    \beq\label{ut}
    \bw^{t} = \tilde{\bY}\tilde{\bv}^t - \frac{n}{d}\tilde{\bu}^{t-1}, 
    \eeq
    \beq 
    \tilde{\bu}^t = g_t(\bw^t), 
    \eeq
    \beq     \label{vt}
 \tilde{\bv}^{t+1} = \tilde{\bY}^\top \tilde{\bu}^t - {\sf c}_t \tilde{\bv}^t, \quad {\sf c}_t = \frac{1}{d}\|\bw^t\|_0,
    \eeq
    where \( g_t(\bw) = \text{sign}(\bw)(|\bw|- \tau_t)_{+} \) is the soft-thresholding operator and the functions are applied entry-wise to \(\bw\).
\end{itemize}

\textbf{Step 3: Renormalization and Output.}
\begin{itemize}
    \item Compute:
    \[
    \widehat{\bv} = \hat{\bF} \tilde{\bv}^L, \quad \widehat{\bu} = (\widehat{u}_1, \ldots, \widehat{u}_d) = \hat{\bH} \tilde{\bu}^L.
    \]
\end{itemize}

\textbf{Output:}{$\widehat{\bv}$, $\widehat{\bu}$.}
\end{algorithm}





Algorithm \ref{alg:iterative_estimation} takes the rescaled data matrix $\tilde \bY$ as input, and initializes $\tilde \bv$ to be the first right singular vector of  $\tilde \bY$. Its key component, Step 2, aims to recovery the rank-one components $\tilde\bu_n$ and $\tilde\bv_n$ from a approximately homeoskedastic matrix $\tilde\bY$. This step is inspired by previous insights that spectral estimation under white noise often yields better performance than under ``colored" noise \citep{leeb2021optimal,gavish2023matrix}. In the iterative AMP process, given $\tilde\bv^t$, we obtain $\bw^t$ by projecting $\tilde \bY$ on $\tilde\bv^t$ subtracting an additional bias correction term $\frac{n}{d}\tilde{\bu}^{t-1}$.  The thresholding operator $g_t(\bw^t)$ represses all the components of $\bw^t$, whose magnitudes are smaller than $\tau_t$, to zero. For some properly chosen thresholding parameter $\tau_t$ (see below for detailed discussions), for sparse vector $\tilde \bu$ and some reasonable estimator $\bw^t$, we expect $g_t(\bw^t)$ to have a smaller angle with $\tilde \bu$ than the original $\bw^t$. As a result, at each step, a better estimator of $\tilde\bu$ will improves the estimation of $\tilde\bv$ through (\ref{vt}), and again leads to a even better estimator of $\tilde\bu$ through (\ref{ut}).  Compared with the standard projected power methods, the additional correction terms $\frac{n}{d}\tilde\bu^{t-1}$ and ${\sf c}_t \tilde\bv^t$ are included in (\ref{ut}) and (\ref{vt}), respectively, which are called  ``Onsager terms." These terms essentially correct the bias introduced at each iteration due to the correlation between $\tilde\bY$ and the estimators obtained from the previous step. The Onsager correction allows for a more refined control of AMP's limiting behaviors in the high-dimensional regime, leading to precise characterization of the estimation accuracy. Once we obtain the estimators for $\tilde\bu_n$ and $\tilde \bv_n$, we remove the rescaling $\hat\bF^{-1}$ and $\hat\bH^{-1}$ due to the variance stabilization, and obtain the final estimators of $\bar\bu$ and $\bv$.


Algorithm \ref{alg:iterative_estimation} requires specifying a sequence of threshold parameters $\{\tau_t\}$ for the soft-thresholding operator to encourage sparsity in $\tilde\bu^t$. Below we discuss a practical data-driven procedure for determining them. 
Intuitively, when the true sparsity $s$ is known, we may consider the following steps for determining $\tau_t$. Suppose $s=\omega d$ for some constant $\omega\in(0,1)$, where $\omega$ is known. Then, at each AMP iteration, we can set $\tau_t$ to be the smallest value satisfying 
\beq
\rho^t_n(\tau_t) \ge \omega, 
\eeq
where
\beq\nonumber
\rho^t_n(x) = \frac{1}{n}\sum_{1\le i\le n} 1\{ |u^t_i|\le x\}.
\eeq
In other words, for known sparsity level $\omega\in (0,1)$, the above procedure would generate a sequence of $\{\tau_t\}_{t\ge L}$ suitable for Algorithm \ref{alg:iterative_estimation}. In particular, our extensive numerical experiments confirm the empirical performance of such a procedure for determining $\{\tau_t\}_{t\ge L}$.

In practice, when the true sparsity  $s$ is unknown, we propose a cross-validation procedure to determine the hyperparameter $\omega$, such that the resulting $\{\tau_t\}_{t \ge L}$ would be similar to those obtained under the oracle setting.

\begin{algorithm}[H]\label{alg:cross_validation}
\caption{Cross-Validation for Optimal $\omega$ Selection}

\textbf{Input:}{$n$ subjects, number of folds $K$, grid of potential $\omega$ values $\Omega$.}

\textbf{Step 1: Split the $n$ subjects into $K$ folds.}

\textbf{Step 2: For each $\omega \in \Omega$:}
\begin{enumerate}
     \item For each fold $k \in [K]$:
    \begin{enumerate}
        \item Learn $\hat{\bu}^{(-k)}(\omega)$ using Algorithms 1 and 2 with $K-1$ folds, leaving the $k$-th fold out.
        \item In the $k$-th fold, normalize the data to obtain $\bar{\bY}^{(k)}$.
        \item Compute:
        \[
        \hat{\bv}^{(k)}(\omega) = \bar{\bY}^{(k)^\top} \hat{\bu}^{(-k)}(\omega).
        \]
        \item Define the loss for the $k$-th fold as:
        \[
        l^{(k)}(\omega) = \|\bar{\bY}^{(k)} - \hat{\bu}^{(-k)}(\omega) \hat{\bv}^{(k)}(\omega)^\top\|^2_2.
        \]
    \end{enumerate}
    End loop over $k$.
\end{enumerate}
End loop over $\omega$.

\textbf{Step 3: Choose the optimal $\hat{\omega}$ as:}
\[
\hat{\omega} = \arg\min_{\omega} \sum_{k=1}^{K} l^{(k)}(\omega).
\]

\textbf{Output:}{Optimal $\hat{\omega}$.}
\end{algorithm}

\section{Theoretical Properties} \label{sec.theory}

\subsection{Consistency of variance stabilization algorithm}

In this part, we establish the consistency of the normalizing matrices $\hat\bH$ and $\hat\bF$ defined in Algorithm \ref{alg:variance_stabilization}. We begin by introducing our theoretical assumptions. In line with our interests in the high-dimensional setting, our analysis will focus on the high-dimensional asymptotic regime. Specifically, we assume

\noindent (A1) the aspect ratio between $n$ and $d$ converges to a deterministic limit $\alpha\in(0,\infty)$, or $d/n\to\alpha.$

For the signal vector $\bv\in\R^n$, we assume 

\noindent (A2) the components of the signal vector $\bv$ are delocalized, that is 
\beq
\|\bv\|_\infty\lesssim \|\bv\|_2\sqrt{\frac{\log n}{n}}.
\eeq

\noindent (A3) the $\ell_2$-norm of $\bv$ converges to some positive constant $\lambda>0$, that is, $\lim_{n\to\infty}\|\bv\|_2=\lambda$.

\noindent In particular, Condition (A2) requires the signal vector to have dense components, a reasonable assumption implying that the underlying risk cannot be exactly zero for a substantial number of subjects, whereas Condition (A3) defines the key parameter $\lambda$ characterizing the overall signal strength in data.

For the sparse indicator vector $\bu\in\R^d$, we assume

\noindent (A4) $u_i\in\{0,\sqrt{\gamma_u/s}\}$ where $\gamma_u<\infty$ is a constant, and the sparsity parameter  $s=\sum_{i=1}^d 1\{u_i\ne 0\}$ satisfies $s/d\to \omega\in(0,1)$.

The parameter $\omega$ characterizes the sparsity level of $\bu$ as well as the normalized vector $\bar\bu$. Moreover, regarding the model-specific noise levels $\{\sigma_i\}_{1\le i\le d}$ and the sample-specific noise levels $\{f_i\}_{1\le i\le n}$, we assume

\noindent (A5) $c<\min_{1\le i\le d}\sigma_i\le \max_{1\le i\le d}\sigma_i<c^{-1}$ and $c<\min_{1\le i\le n}f_i\le \max_{1\le i\le n} f_i<c^{-1}$ for some universal constant $0<c<1$.

Next we define the convergence limits of the normalizing matrices $\hat\bH$ and $\hat\bF$, respectively. As we noted in Equation (\ref{eq:S_rank1}), the the variance of the heteroskedastic noise $\bE$ has an approximate rank one structure with $\bS = \frac{1}{n}\bh_0\fb^\top$.  
However,  $\bh_0$ and $\bbf$ are  identifiable in $\bS$ only up to a global rescaling factor. Here, we show the convergence of $\hat\bH$ and $\hat\bF$ with respect to some rescaled versions of $\bh_0$ and $\bbf$, defined as $\bar\bh=(\bar h_1, \bar h_2, ..., \bar h_d)$ and $\bar\bbf=(\bar f_1, \bar f_2, ..., \bar f_n)$, satisfying 
\beq\label{hf.bar1}
\bS=\bar\bh\bar\bbf
\eeq
 and
  \beq \label{dyson}
\bar\bh = \frac{1}{\sqrt{d-\eta\|\bg^{(1)}\|_1}}\bigg(\frac{1}{\bg^{(1)}} -\eta\bigg),\qquad \bar\fb = \frac{1}{\sqrt{n-\eta\|\bg^{(2)}\|_1}}\bigg(\frac{1}{\bg^{(2)}} -\eta\bigg),\qquad \frac{\bar\bh^\top\bg^{(1)}}{\bar\fb^\top\bg^{(2)}}=1,
\eeq
where $\eta$ is the deterministic limit of $\bar\theta$ \citep{erdHos2019random}.

It is easy to show that there exist $a,b\in\R_{+}$ that $\bar\bh=a\bh_0$ and $\bar\fb=b\bbf$. As a result, $\bar\bh$ and $\bar\fb$, by definition, are simply rescaled versions of $\bh_0$ and $\fb$, respectively. We define
\[
\bar\bH = \text{diag}(n^{1/4}\bar\bh^{1/2}),\qquad \bar\bF = \text{diag}(n^{1/4}\bar\bbf^{1/2}).
\]

Our next theorem concerns the rates of convergence of $\hat\bH$ and $\hat\bF$ to $\bar\bH$ and $\bar\bF$ as $(n,d)\to\infty$.

\begin{thm} \label{g.bnd.prop}
Suppose the predicted scores $\bY_i\in\R^n$ are generated from the model (\ref{datamodel}). Under Assumptions (A1)-(A5), for any $\epsilon>0$, we have
\beq \label{HF.conv}
\|\hat\bH-\bar\bH\|\le Cd^{-1/2+\epsilon},\qquad \|\hat\bF-\bar\bF\|\le Cd^{-1/2+\epsilon},
\eeq
with probability at least $1-d^{-c}$ for some constants $C, c>0$.
\end{thm}

Theorem \ref{g.bnd.prop} justifies the use of  Algorithm \ref{alg:variance_stabilization} for variance stabilization, showing that the two diagonal matrices $\hat\bH$ and $\hat\bF$ can effectively correct for the noise heteroskedasticity in (\ref{norm.eq}). Interestingly, our analysis of the rates of convergence in (\ref{HF.conv}) suggests that, despite the fact that combining multiple pre-trained models introduces heteroskedasticity, our method will benefit from a larger number of pre-trained models  and a larger number of samples, by leveraging the approximate rank-one structure embedded in their underlying noise structures.

\subsection{Asymptotic limits of approximate message passing estimators} \label{sec.amp.theory}

We begin by introducing some notations and definitions needed to present our main theoretical results.

\begin{defn}
   (weak convergence) A sequence of probability distributions $\nu_n$ on $\R^m$ converges weakly to $\nu$, or $\nu_n\to_w\nu$, if for any bounded Lipschitz function $\psi:\R^m\to\R$, we have $\lim_{n\to\infty}\E\psi(X_n) =\E \psi(X)$ where $X_n\sim \nu_n$ and $X\sim \nu$. In particular, for a (deterministic) sequence of vectors $\{Z_i\}_{1\le i\le n}$ in $\R^m$, we say its empirical distribution converges weakly to $\nu$ on $\R^m$ if $\lim_{n\to\infty}n^{-1}\sum_{i=1}^n\psi(Z_i)=\E\psi(Z)$ for any bounded Lipschitz function $\psi$, where $Z\sim \nu$.
\end{defn}

Next we introduce some additional assumptions needed for our theoretical analyses. Specifically, regarding the signal vector $\bv\in\R^n$, we assume

\noindent (A6) the empirical distributions of the  rescaled entries $\{\lambda^{-1}\sqrt{n}v_i\}_{1\le i\le n}$ of $\bv$ converge weakly to a probability distribution $\nu_v$ on $\R$ with unit second moment.

Note that the unit second moment condition is consistent with our Assumption  (A3) regarding the $\ell_2$-norm of $\bv$. Moreover, after variance stabilization, from the expression (\ref{Ytilde}) and the relationships (\ref{ubar}) and (\ref{HF.conv}), we can see that the signal component of $\tilde\bY$ is approximately $\bar\bH^{-1}\bar\bu_0\bv^\top\bar\bF^{-1}$, where 
\[
\bar\bu_0 = \bigg(\frac{u_1}{\sqrt{\lambda^2u_1^2+\sigma^2_1}}, ..., \frac{u_d}{\sqrt{\lambda^2u_d^2+\sigma^2_d}}\bigg)^\top.
\]
In particular, after variance stabilization, the two components of the rank-one signal matrix becomes
\beq
\tilde\bu=(\tilde u_1,...,\tilde u_d) = \bar\bH^{-1}\bar\bu_0,\qquad \tilde\bv = (\tilde v_1,...,\tilde v_n)= \bar\bF^{-1}\bv.
\eeq
In particular,
for each $i\in\{1,...,d\}$, the $i$th row of $\tilde\bY$ satisfies $\E \tilde\bY_i\approx \tilde u_i\cdot \tilde \bv$, where
\beq\label{beta}
\tilde u_i=\frac{u_i}{n^{1/4}\bar h_i^{1/2}\sqrt{\lambda^2u_i^2+\sigma_i^2}},
\eeq
and
\beq
\tilde v_j=n^{-1/4}v_j \bar f_j^{-1/2}.
\eeq
In other words, after variance stabilization, the relative signal strength contained in each $\tilde\bY_i$ is captured by the weights $\{\tilde u_i\}_{1\le i\le d}$, and the normalized signal vector $\bv$ becomes $\tilde\bv$. 
Note that $\{\tilde u_i\}_{1\le i\le d}$ depends on $\{u_i\}_{1\le i\le d}$ $\{\sigma_i\}_{1\le i\le d}$ and $\lambda$, whereas $\{\tilde v_j\}_{1\le j\le n}$ depend on $\{v_j\}_{1\le j\le n}$ and $\{\bar f_j\}_{1\le j\le n}$. Our next two assumptions essentially calibrate the scaling of the model-specific weights $\{\tilde u_i\}_{1\le i\le d}$ and the sample-specific signals $\{\tilde v_j\}_{1\le j\le n}$, so that the parameter $\lambda$ will capture the global signal-to-noise ratio of the model. Specifically, regarding the model-specific weights $\tilde\bu$, we assume

\noindent (A7) the empirical distributions of $\{\sqrt{n}\tilde u_i\}_{1\le i\le d}$ converge weakly to a probability distribution $\nu_u^*$ on $\R$, with unit second moment.

The unit second moment condition in (A7) ensures that the values in $\{\tilde u_i\}$ are properly scaled so that  the norm $\|\bar\bH^{-1}\bar\bu_0\|_2^2=\sum_{i=1}^d\tilde u_i^2$ converges to 1. Similarly, regarding the normalized sample-specific signals $\tilde\bv$, we assume

\noindent (A8) the empirical distributions of $\{\sqrt{n}\tilde v_j/\sqrt{\lambda}\}_{1\le j\le n}$ converge weakly to a probability distribution $\nu_v^*$ on $\R$, with unit second moment.

The unit second moment condition in (A8)  ensures that the values in $\{\bar f_j\}_{1\le j\le n}$ are properly normalized so that they only reflect the cross-sample noise variations in the relative sense, so that the norm of $\tilde\bv$ remains the same as the norm of $\bv$, which converges to $\lambda$. The two assumptions (A7) and (A8), combined together, ensures that the absolute SNR of $\tilde\bY$ is captured by $\lambda$ alone. In particular, Assumptions (A7) and (A8) are satisfied by diverse heteroskedastic noise scenarios for $\{\sigma_i\}_{1\le i\le n}$ and $\{f_j\}_{1\le j\le n}$. See Section \ref{sup.sec.theory} in the Supplement for additional discussions.

Conditions (A6)-(A8) relate the finite-sample distribution of the components of $\{v_i\}$, $\{\beta_i\}$
and $\{f_i\}$ to some well-defined populations, characterizing their global properties. These assumptions essentially only concern the suitable scaling of various model parameters, to ensure their respective identifiability, the existence of the asymptotic limits as expressed in (\ref{amp.lim}), and the interpretation of $\lambda$ as the global signal-to-noise ratio. They are standard technical assumption required by rigorous analysis of AMP algorithms \citep{donoho2009message,bayati2011dynamics}.
Below we state our main theorem concerning the convergence of the AMP algorithm.

\begin{thm} \label{main.thm}
Suppose Assumptions (A1)-(A7) hold and $\lambda^2\sqrt{\alpha}>1$. Let $(\mu_t,\sigma_t)_{t\ge 0}$ be defined via the recursion
\beq\label{rec1}
\mu_{t+1} = \lambda \bar\mu_t, \quad \sigma_{t+1}^2 =\bar\mu_t^2+ \bar\sigma_t^2,
\eeq
\beq \label{rec2}
\bar{\mu}_t = \lambda \alpha \, \mathbb{E} [U g_t(\mu_t U + \sigma_t G)], \quad \bar{\sigma}_t^2 = \alpha \, \mathbb{E} [g_t(\mu_t U + \sigma_t G)^2],
\eeq
where \( U \sim \nu_{u}^* \),  and \( G \sim \mathcal{N}(0,1) \) are independent, and the initial condition is
\[
\mu_0 = \sqrt{\frac{1 - \alpha^{-1} \lambda^{-4}}{1 + \lambda^{-2}}}, \quad \sigma_0 = \sqrt{\frac{\lambda^{-2} + \alpha^{-1} \lambda^{-4}}{1 + \lambda^{-2}}}.
\]
(This is to be substituted in Eq. (\ref{rec2}) to yield \( \overline{\mu}_0, \overline{\sigma}_0 \).)
Suppose that $\tilde\bv^\top\hat\bv\ge 0$.
Then, the following holds almost surely for \( t \geq 0 \):
\beq \label{amp.lim}
\lim_{n\to\infty}\frac{\langle \tilde\bv,\tilde\bv^t\rangle}{\|\tilde\bv\|_2\|\tilde\bv^t\|_2} =  \frac{\mu_{t+1}}{\lambda\sigma_{t+1}}, 
\qquad
\lim_{n\to\infty} \frac{\langle\tilde\bu,\tilde\bu^t\rangle}{\|\tilde\bu\|_2\|\tilde\bu^t\|_2} = \frac{\bar\mu_t}{\sqrt{\alpha}\lambda\bar\sigma_t}.
\eeq
\end{thm}

Theorem \ref{main.thm} provides high-dimensional asymptotic limits for the normalized cosine similarity between the AMP estimators at each given $t\ge 0$ and their objectives $(\tilde\bu, \tilde\bv)$, which are modified versions of $(\bar\bu_0,\bv)$ after variance stabilization. Combining Theorems \ref{g.bnd.prop} and \ref{main.thm}, it follows that whenever the AMP estimators $(\tilde\bu^L, \tilde\bv^L)$ are sufficiently close to $(\tilde\bu, \tilde\bv)$, the final  estimators $(\widehat\bu, \widehat\bv)=(\hat\bH\tilde\bu^L,\hat\bF\tilde\bv^L)$ will be necessarily close to the parameters of interest $(\bar\bu_0,\bv)$. 

A few remarks of Theorem \ref{main.thm} are in order. First, the requirement $\lambda^2\sqrt{\alpha}>1$ is the minimum signal-to-noise ratio condition necessary for the estimation of the singular vectors from noisy observations; in random matrix theory, such a requirement is also characterized as the BBP (Baik-Ben Arous-Peche) phase transition \citep{johnstone2001distribution,baik2005phase,bao2021singular}.  Second, although in this study we focus on the soft-thresholding functions $g_t(\cdot)$ to encourage sparsity in $\tilde\bu^t$, from our proof  it can be seen that the above theorem holds for any separable Lipschitz function \citep{montanari2021estimation}. Thirdly, regarding the interpretation of $\lambda$, by Assumptions (A3) and (A6)-(A8), it can be seen that, in practice, a larger value of $\lambda$ could be a consequence of an overall better predictive accuracy of the informative models (with $u_i\ne 0$), or a smaller number of noninformative models (with $u_i=0$). 

\begin{figure}
    \centering
\includegraphics[width=0.8\linewidth]{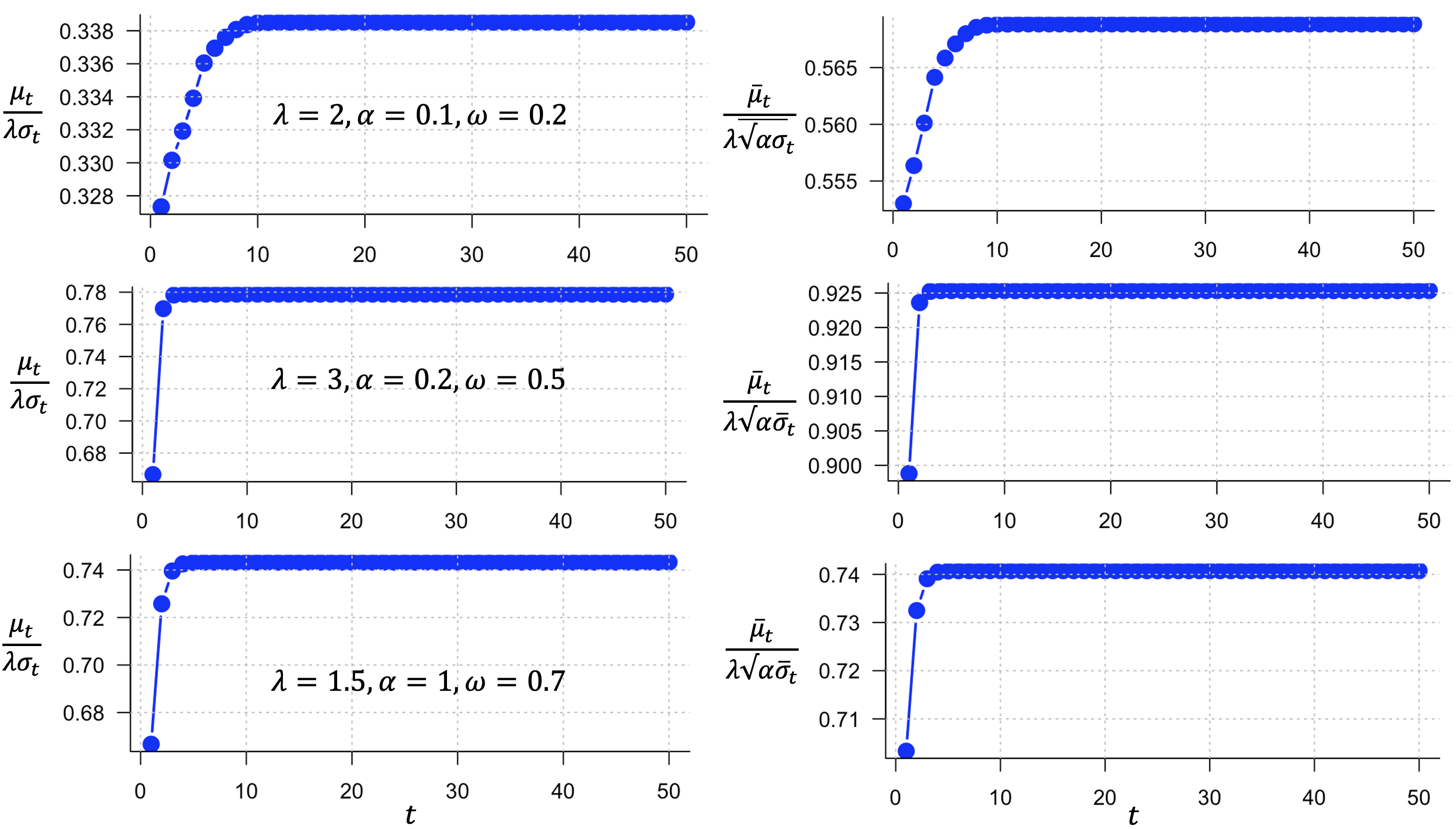}
    \caption{Values of $\mu_t/(\lambda\sigma_{t+1})$ (Left) and $\bar\mu_t/(\sqrt{\alpha}\lambda\sigma_{t+1})$ (Right) of the AMP algorithm over iteration numbers $t$. The specifications of $\lambda, \alpha$, and $\omega$ are shown in the plot. $\nu_u^*$ is a mixture distribution, consisting of a uniform distribution over $(0, c)$ with probability $\omega$, and a point mass at $0$ with probability $1 - \omega$. We choose $c$ such that the second moment of $\nu_u^*$ is $1$. The two plots on the same row have the same parameter setting.}
    \label{fig:theorem1}
\end{figure}

The asymptotic limits in (\ref{amp.lim}) are expressed in terms of a  sequence $\{(\mu_t, \sigma_t,\bar\mu_t,\bar\sigma_t)\}_{t\ge 0}$ of parameters defined by the recursive formulas given by (\ref{rec1}) and (\ref{rec2}). These parameters are uniquely determined by  the model parameters $(\lambda, \alpha, \nu_u^*, \nu_v^*)$ and the thresholding function $g_t(\cdot)$. To better illustrate the implications of (\ref{amp.lim}), we consider some specific choices of $(\lambda, \alpha, \nu_\beta, \nu_\theta)$ and a fixed thresholding function $g_t(\bw)\equiv g(\bw)$ for some thresholding parameter, and present in Figure \ref{fig:theorem1} the numerical values of the asymptotic limits $\mu_{t+1}/(\lambda \sigma_{t+1})$ and $\bar\mu_t/(\sqrt{\alpha}\lambda\bar\sigma_t)$, as a function of $t$. We observe that these values first increase in $t$, and then become stationary, suggesting a convergence of these sequences themselves.

In this case, for sufficiently large $t$, the parameters $(\mu_t, \sigma_t,\bar\mu_t,\bar\sigma_t)$ converge to the fixed point $(\mu^*, \sigma^*,\bar\mu^*,\bar\sigma^*)$ of the recursion (\ref{rec1}) and (\ref{rec2}), which is also the solution to the following system of nonlinear equations
\beq
\begin{cases}\mu^* = \lambda \bar\mu^* \\
{\sigma^*}^2 ={\bar{\mu^*}}^2+ \bar{\sigma^*}^2\\
\bar{\mu}^*= \lambda \alpha \, \mathbb{E} [U g(\mu^* U + \sigma^* G)]\\ 
\bar{\sigma^*}^2 = \alpha \, \mathbb{E} [g(\mu^* U + \sigma^* G)^2]
\end{cases}.
\eeq
In other words, with  large iterations, the asymptotic limits are
\beq 
\lim_{t\to\infty}\lim_{n\to\infty}\frac{\langle \tilde\bv,\tilde\bv^t\rangle}{\|\tilde\bv\|_2\|\tilde\bv^t\|_2} =  \frac{\mu^*}{\lambda\sigma^*}, 
\qquad
\lim_{t\to\infty}\lim_{n\to\infty} \frac{\langle\tilde\bu,\tilde\bu^t\rangle}{\|\tilde\bu\|_2\|\tilde\bu^t\|_2} = \frac{\bar\mu^*}{\sqrt{\alpha}\lambda\bar\sigma^*}.
\eeq

\begin{figure}
    \centering
\includegraphics[width=0.5\linewidth]{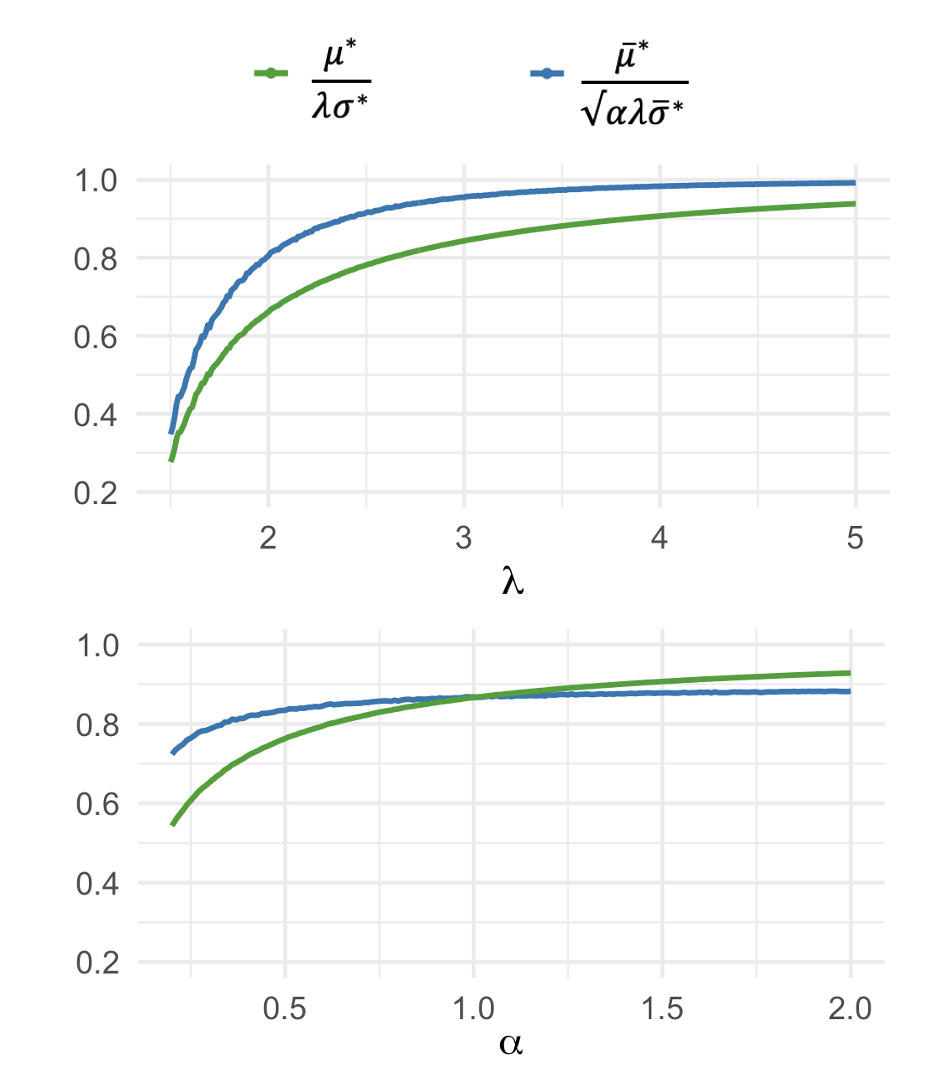}
    \caption{Illustration of the asymptotic limits $\mu^*/(\lambda\sigma^*)$ (Green) and $\bar\mu^*/(\sqrt{\alpha}\lambda\sigma^*)$ (Blue) as functions of $\lambda$ (Upper Panel) and $\alpha$ (Lower Panel). In the upper panel, we choose $\alpha = 0.3$ and $\omega = 0.3$. In the lower panel, we choose $\lambda = 2$, and $\omega = 0.5$. In all settings, $\nu_u^*$ is a mixture distribution, consisting of a uniform distribution over $(0, c)$ with probability $\omega$, and a point mass at $0$ with probability $1 - \omega$. We choose $c$ such that the second moment of $\nu_u^*$ is $1$. }
    \label{fig:theorem2}
\end{figure}

{Although a closed-form expression is not available, Figure~\ref{fig:theorem2} presents the numerical values of $\frac{\mu^*}{\lambda\sigma^*}$ and $\frac{\bar\mu^*}{\sqrt{\alpha}\lambda\bar\sigma^*}$ as functions of $\lambda$ and $\alpha$, respectively, under representative settings.
}We observe that as $\lambda$ or $\alpha$ increases, both limits $\frac{\mu^*}{\lambda\sigma^*}$ and $\frac{\bar\mu^*}{\sqrt{\alpha}\lambda\bar\sigma^*}$  increase. This demonstrate  the consistency of these AMP estimators. In sum, our analysis suggests that our proposed AMP algorithm, and therefore our final consensus risk score prediction $\widehat\bv$, will achieve better predictive accuracy if one starts with a collection of pre-trained models with better overall predictive accuracy (bigger $\lambda$), or if there are a larger numbers of prediction models (bigger $\alpha$). 

\section{Simulation Study} \label{sec.simu}

We conduct simulations to evaluate the performance of the proposed method in comparison to several benchmark approaches for unsupervised model aggregation or low rank signal estimation. Data is generated according to equation (\ref{datamodel}). Specifically, the true signal vector $\bv\in\R^n$ is sampled from a uniform distribution over the interval $(-1, 1)$. We randomly draw each $u_i$ from $\{0, \sqrt{\alpha/s}\}$ where $\alpha = d/n$ varies across settings. The sparsity level $s$ is controlled such that the ratio $\omega = s/d$ is fixed at values of 0.1, 0.3, 0.5, and 0.7. We generate the subject-specific noise vector $\fb\in\R^n$  and the model-specific parameter $\sigma_i$ from certain distributions which vary across simulation settings.
The random noise $\bw_i$ is generated as $\bw_i \sim N(0, \frac{1}{n}{\bf I})$. We set the number of samples $n =1000$, and let the number of models $d$ vary from $50$ to $200$. 

We consider two versions of the proposed method, where one assumes the sparsity  ratio $\omega$ is known (U-aggregation-o) and the other uses the cross validation procedure proposed in Algorithm 3 to learn $\omega$ (U-aggregation) from the grid $\Omega = (0.1,0.2, \dots, 0.9)$. We also consider  the following alternative methods in our simulation: 
\begin{enumerate}
    \item Simple average: we normalize $\bY_i$ to get $\bar \bY_i$ and take the average  $\frac{1}{d}\sum_{i=1}^{d}\bY_i$.
    \item PCA: we apply principal component analysis on $G = \bar \bY_i \bar \bY_i^\top \in \R^{d\times d}$, and use the first eigen vector estimate $\bv$.
    \item HeteroPCA: we apply the heteroPCA algorithm proposed by \cite{zhang2022heteroskedastic} on $G = \bar \bY_i \bar \bY_i^\top \in \R^{d\times d}$ and use the first eigenvector estimate $\bv$.
\end{enumerate}
We consider two settings. In the homoskedastic setting we generate $\fb_j = 1$ for all $i\in[n]$ and the model-specific parameter $\sigma_i=1$ all $i\in[d]$. In the heteroskedastic setting, we sample $\fb_j$ and $\sigma_i$ both from a uniform distribution over $(0, 2)$.  Note that in the above simulation settings, we choose not to scale $\bu$, $\bf$, and $\bh$ according to  Assumption (A6)-(A8), which are imposed to facilitate theoretical analysis. However, the mere change of parameter scaling will not affect the interpretation of our results.
\begin{figure}
    \centering
\includegraphics[width=1\linewidth]{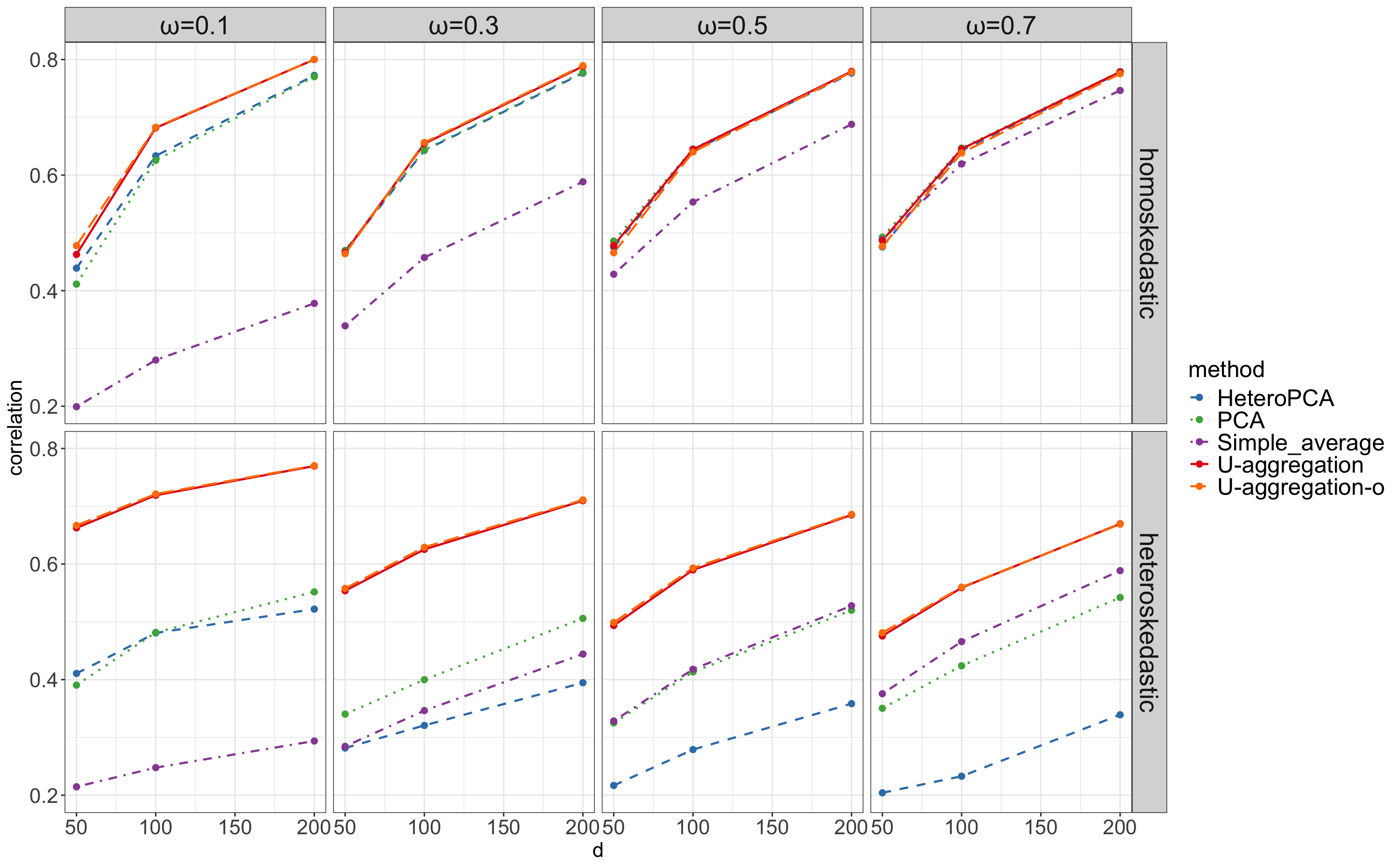}
    \caption{Performance of the compared methods in estimating $\bv$ across various simulation settings.}
    \label{fig:simu_correlation}
\end{figure}

Figure \ref{fig:simu_correlation} shows the performance of all the compared methods, evaluated by the Person's correlation between $\bv$ and its estimates from the above list of methods across all simulation settings.  We first observe that the U-aggregation performs nearly the same as if we know the true sparsity, and Figure S1 in the Supplementary Material shows the estimated $\omega$ against the truth. We can see the the estimation is fairly good especially when $\omega$ is small. These demonstrate the effectiveness of the CV approach. Comparing to other methods, with  homoskedastic noises,  U-aggregation outperforms  HeteroPCA and PCA  when $\omega$ is less than 0.3 and they perform equaly well when  $\omega$ is large. As $\omega$ increases, the simple average method which assigns equal weights to all models has performance closer to PCA and U-aggregation. On the other hand, when having heteroskedastic noises, U-aggregation performs much better than all the alternative approaches. The HeteroPCA seems to be not effective under our data generating model, performing even poorly than PCA.

\begin{figure}
    \centering
    \includegraphics[width=1\linewidth]{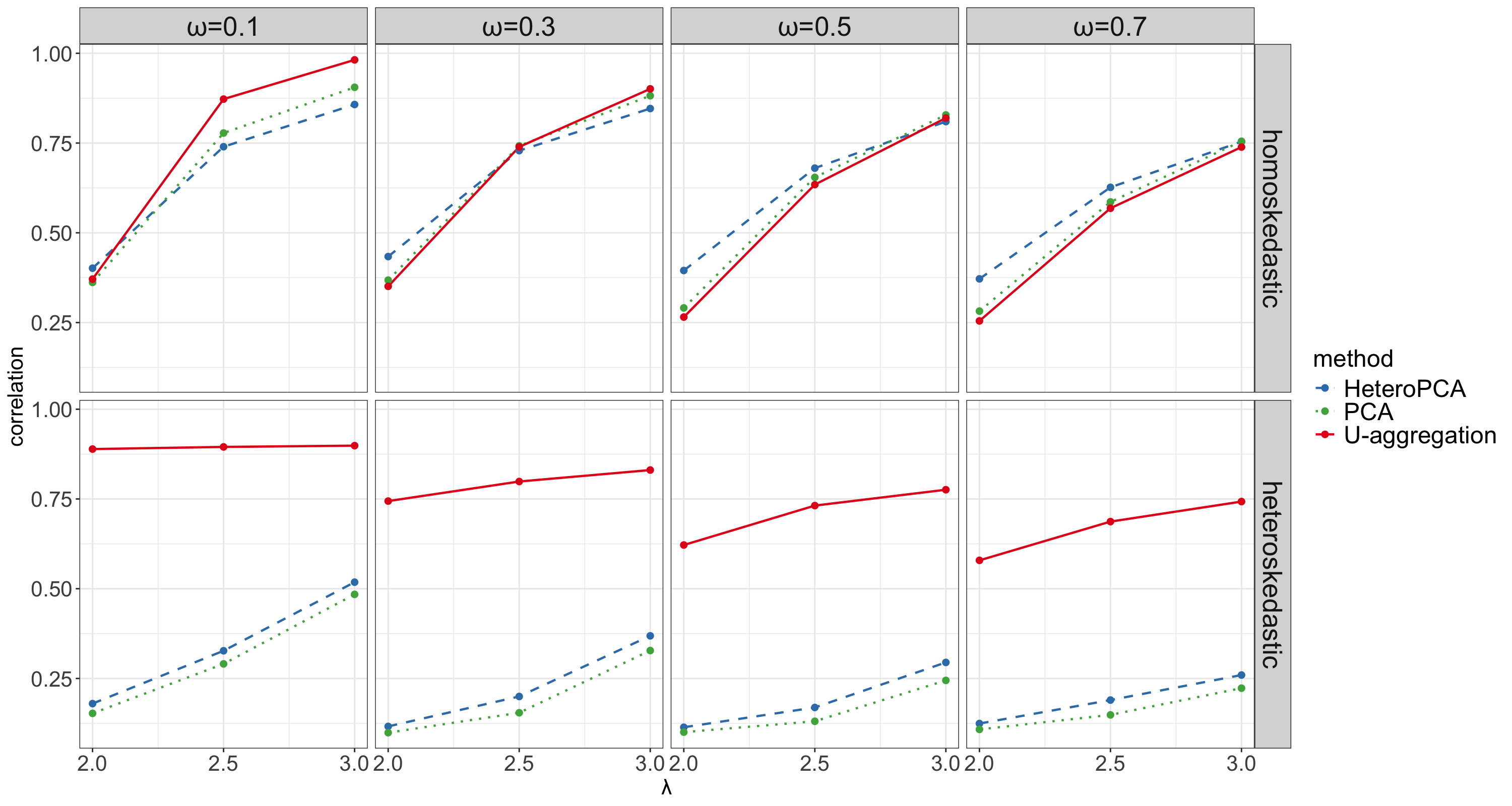}
    \caption{Correlation between the assigned weights and the performance of each pre-trained model across various simulation settings.}
    \label{fig:u}
\end{figure}

We additionally evaluated how well the estimated $\hat \bu$ can characterize the performance of each candidate model. Instead of directly comparing $\hat\bu$ to $\bu$ or $\bar\bu$, a more relevant evaluation in practice is to compare $\hat u_i$ with the actual performance of each model. We therefore quantify the performance of each model by $\rho_i = \text{cor}(\bY_i,\bv)^2$, and obtain $\bf{\rho} = (\rho_1,\rho_2,\dots,\rho_d)$. We then quantify the concordance between $\hat \bu$ and $\bf{\rho}$ by their corelation, as shown in Figure \ref{fig:u}. We compare U-aggregation with PCA and HeteroPCA. Simple averaging is not included in this comparison as it is not designed to incorporate this feature.  For this evaluation, we fix $d = 100$, and $n=1000$, and  increase $\lambda = \|\bv\|_2$. As the signal strength increases, all methods have increased performance for evaluating the performance of the pre-trained models. Similar as the results shown for estimating $\bv$, U-aggregation shows much higher effectiveness  under the heteroskedastic settings, while performing comparable with PC under homoskedastic settings. A. We also observe that HeteroPCA perform slightly better than PCA in various settings.

\section{Integrating pre-trained PRS models to a diverse new population}

In this section, we demonstrate a potential application of our method: integrating open-source pre-trained models from the PGS Catalog to perform genetic risk profiling in a new population. PRS predict an individual’s genetic predisposition to specific traits by aggregating numerous genetic variants identified in genome-wide association studies (GWAS) \citep{khera2018genome, kullo2022polygenic}. While PRSs are valuable for risk stratification and early identification of high-risk individuals, their performance varies across populations and often lacks transferability to genetically diverse groups \citep{wojcik2019genetic, duncan2019analysis, fatumo2022roadmap}. Retraining or calibrating PRS models for new populations can enhance accuracy, but it is often impractical due to limited access to GWAS data, linkage disequilibrium (LD) matrices, and potential computation and human resources required for model training \citep{wray2007prediction, martin2017human}. The PGS Catalog \citep{lambert2021polygenic} provides open-source, well-annotated PRS models, which becomes a valuable resource for direct clinical implementation of pre-trained PRS models.  However, for any given trait, the PGS Catalog often includes multiple models with varying performance in the target population. Additionally, documentation errors may occur when models are uploaded to the PGS Catalog, which could affect model functionality or reliability. We envision that our method can be extremely helpful for practitioners  obtain a robust PRS score for a target population, without the need of curating disease status to retrain the model.

To demonstrate the idea, we draw our target population from the AoU Research Hub, a large, diverse U.S.-based cohort supporting biomedical research to improve health outcomes \citep{all2019all}. AoU enrolls over one million participants nationwide, gathering extensive health data—including whole genome sequencing, genotyping arrays, electronic health records, and survey responses—enabling researchers to analyze the interactions between genetics, environment, and lifestyle factors in health and disease. 

We applied unsupervised aggregation methods to four commonly measured complex continuous traits: height, body mass index (BMI), high-density lipoprotein (HDL) cholesterol, and low-density lipoprotein (LDL) cholesterol, associated with trait IDs EFO\_0004339, EFO\_0004340, EFO\_0004611, and EFO\_0004612, respectively. To evaluate the performance of these methods, we extracted the observed trait data from the AoU research program and excluded patients under the age of 21 at enrollment.  For patients with repeated measurements, we calculated the mean of all observed values for each trait. We then downloaded all available PRS models from the PGS Catalog corresponding to each trait and performed scoring on all genotyped patients. Some PGS models do not contain the chromosome, position, and alleles for each SNP; thus, they were not scored and were removed from the analysis. {In total, 85 models for height, 92 models for BMI, 84 models for HDL, and 97 models for LDL were analyzed.} Meta information regarding these models are included in the Supplementary Table 1. We set the sample size for our target population at 3,000, mimicking the size of a moderate study. Samples were randomly drawn from AoU. 

\subsection{Model Checking}\label{sec.model}
\begin{figure} 
    \centering
\includegraphics[width=1\linewidth]{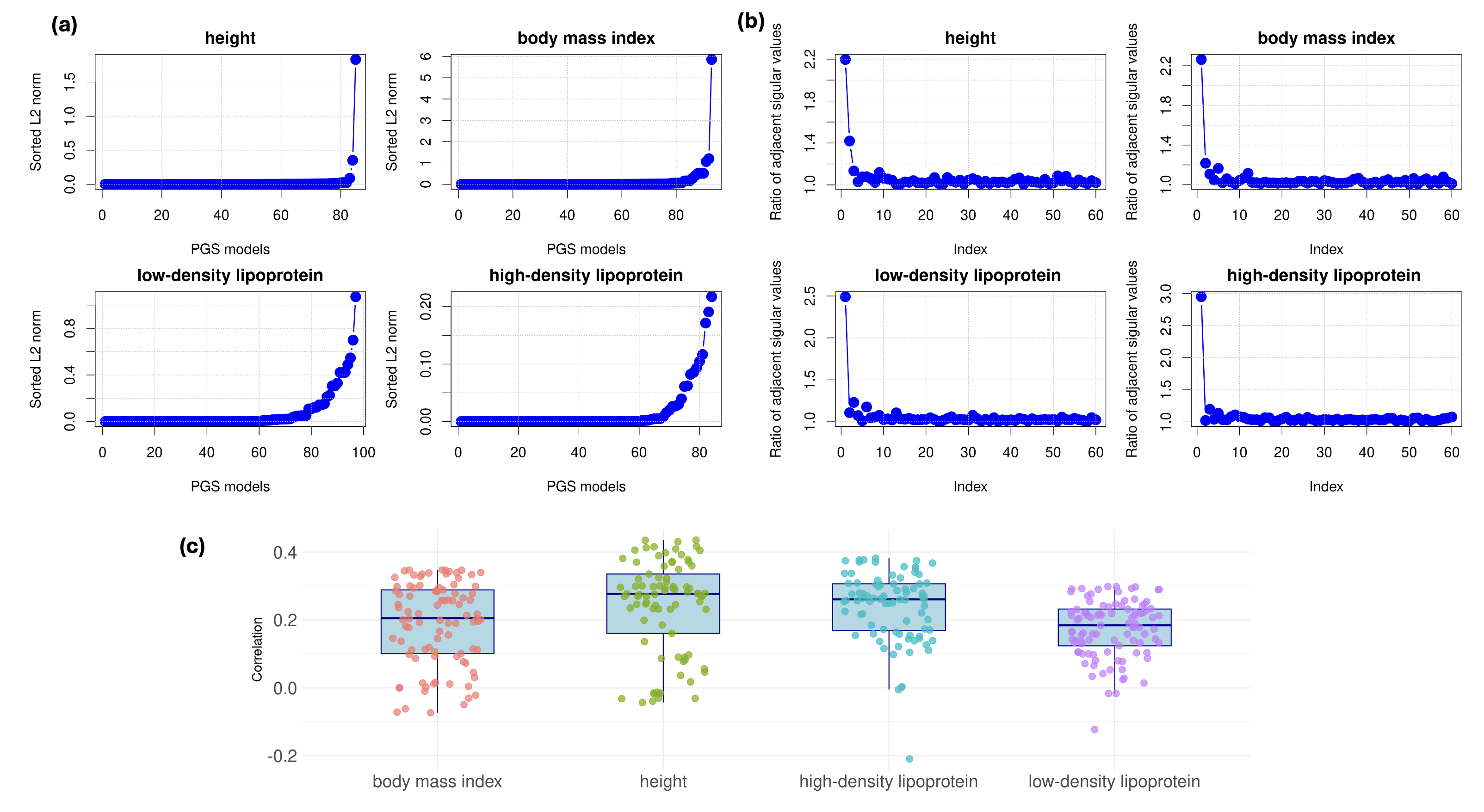}
    \caption{The ratios between consecutive singular values in four datasets from the AoU Study, each containing the pretrained PRSs for a specific trait. It suggests approximate rank-one structures in these matrices.}
     \label{fig:rc}
\end{figure}

Prior to normalization, we observe scaling differences among models for the same trait, as reflected by the $\|\bY_i\|_2$ values after centering, shown in Figure~\ref{fig:rc} (a). For each trait, a small subset of models exhibits significantly larger magnitudes of predicted values compared to the majority, making them less directly comparable. This observation motivates the introduction of a global scaling factor, $c_i$, in the model (\ref{datamodel}). After  normalizing the predicted values from all pre-trained models for each trait (obtaining $\bar\bY$), Figure \ref{fig:rc} (b) illustrates the ratio of consecutive singular values across four datasets from the AoU Study, each corresponding to pre-trained PRSs for a specific trait. These ratios indicate an approximate rank-one structure in the matrices, aligning with our model setup.
The model-specific heteroskedasticity can be partially assessed by evaluating the performance of each model, measured through the correlation between predicted risks and observed trait values. In our data, these correlations range from $-0.1$ to $0.4$, as illustrated in Figure \ref{fig:rc} (c). Many PRS models has correlation near $0$, which are essentially not predictive of the trait of interest. Subject-level heterogeneity is challenging to directly evaluate; however, it is reasonable to assume in practice that the risk for certain individuals may be more difficult to accurately capture compared to others.

\subsection{Comparison with existing model aggregation methods}

We applied various aggregation methods, including U-aggregation, simple average, PCA, and HeteroPCA, to obtain aggregated PRS scores for each trait. We then assessed each method by calculating the correlation between the aggregated PRS scores and the actual values of the continuous traits. To assess the variability of the results, we conducted repeated sampling and get 50 replications in total. 

\begin{figure}
    \centering
    \includegraphics[width=0.9\linewidth]{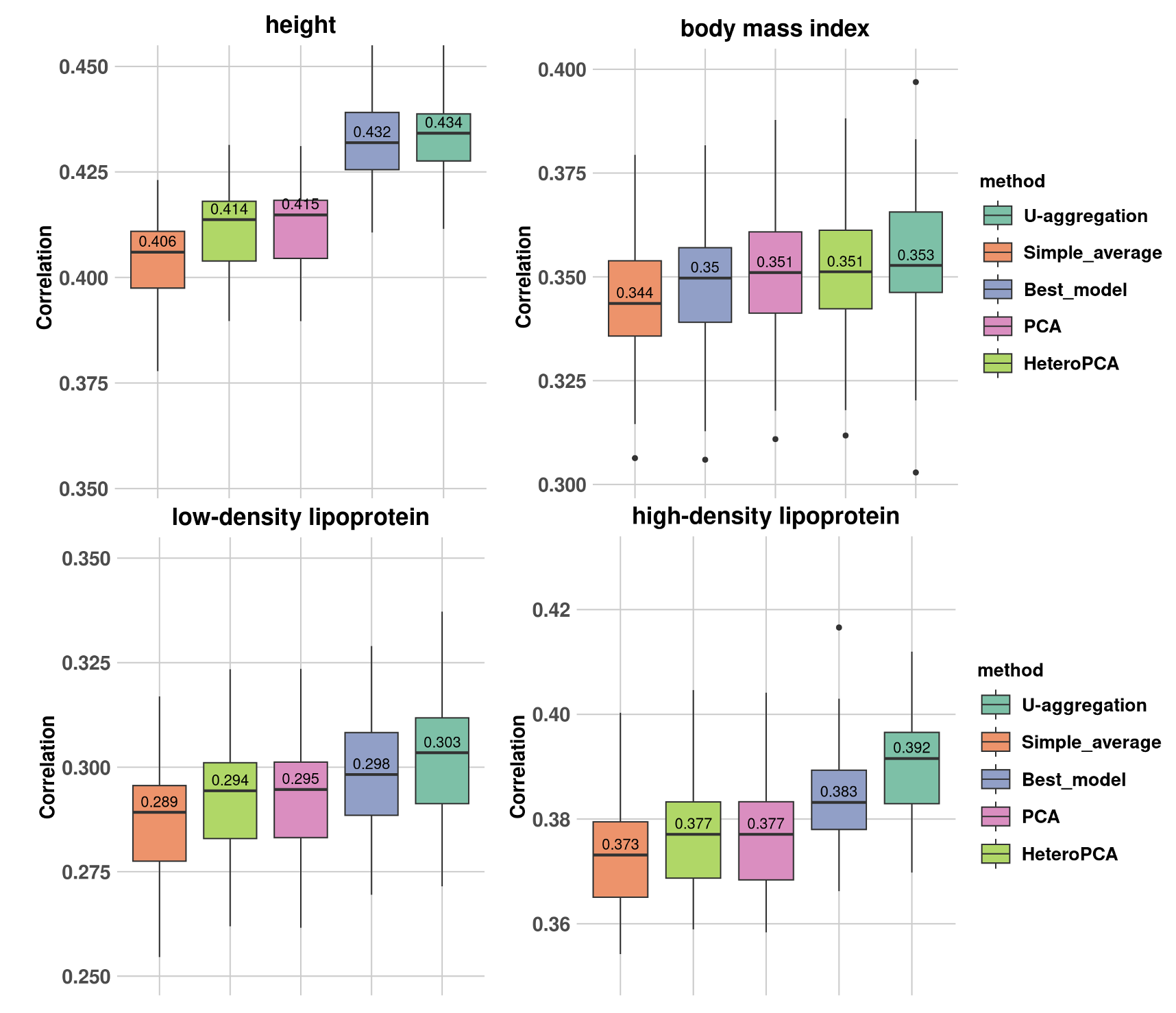}
    \caption{Correlation between the predicted values of each methods and the observed  trait measure. Each box is generated from $50$ random draws from the AoU database.}
    \label{fig:data}
\end{figure}

\begin{figure}
    \centering
    \includegraphics[width=0.6\linewidth]{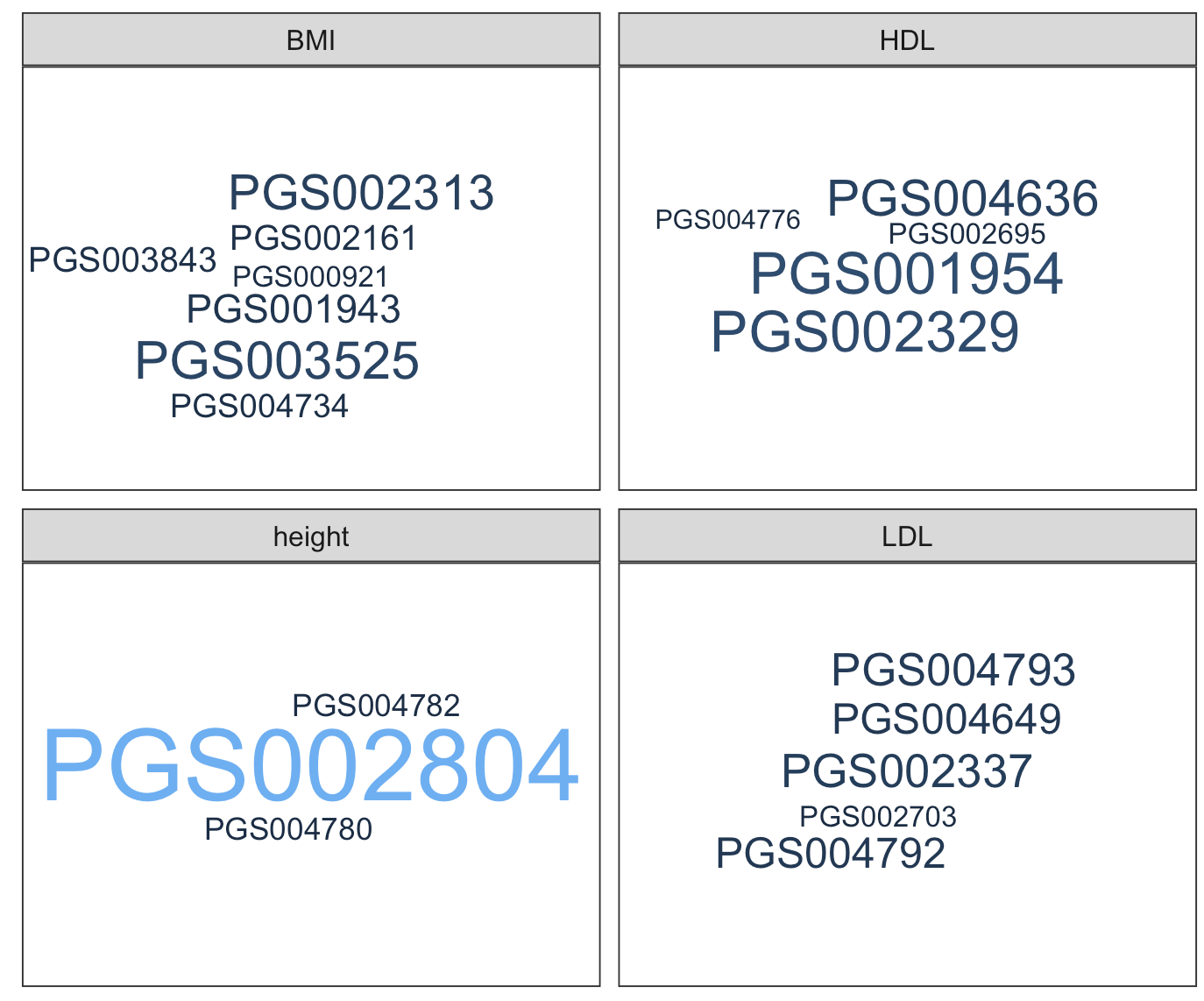}
    \caption{Wordclouds showing the frequency of PRS models being the best model over the $50$ random draws from the AoU database.}
    \label{fig:wordcloud}
\end{figure}

For each target sample, we also identify the pre-trained model that achieves the highest correlation with the true observed trait, denoting it as the "best model" in Figure \ref{fig:data}. Remarkably, U-aggregation slightly outperforms this best model across all traits, despite not using the true outcome to determine the weights. Additionally, the "best model" selected by each sample varies across the 50 random replications, as illustrated in the word cloud in Figure \ref{fig:wordcloud}, which highlights the frequencies of models being selected as the best. In contrast, U-aggregation remains robust to changes in the best model and serves as a reliable alternative in the absence of true outcomes.

The correlation between  weights assigned by U-aggregation to each pre-trained model and their true model performance ($R^2$ of each model) is high, ranging from \(0.86\) to \(0.91\) for height, \(0.79\) to \(0.87\) for BMI, \(0.73\) to \(0.80\) for HDL, and \(0.75\) to \(0.82\) for LDL. These results highlight the potential use of U-aggregation to rank methods effectively, even in the absence of true outcome data. Performance scores offer valuable insights into the quality of PRS models and the methods used to develop them. Sharing these scores with the PGS Catalog could enhance model documentation, improve transparency, and guide future applications.

\section{Discussion}

In this paper, we propose an unsupervised model aggregation approach to derive aggregated predictions by combining multiple pre-trained models. Unlike existing unsupervised aggregation methods, which primarily rely on spectrum decomposition of the fitted value matrix, our approach accounts for potential adversarial models, patient- and model-specific heteroskedasticity, and extends to include the homoskedastic model underlying traditional methods as a special case. We proposed the U-aggregation procedure which involves  a variance stabilization and an iterative sparse rank-one signal recovery algorithm via approximate message passing. In simulation studies, U-aggregation performs similarly to the existing unsupervised model aggregation methods, and significantly outperforms them in the existence of heteroskedasticity. We applied U-aggregation to integrate pre-trained PRS models for height, BMI, LDL, and HDL in the AoU population. U-aggregation demonstrated superior performance, outperforming all existing unsupervised model aggregation methods and even surpassing the best model selected based on the true outcome across all four traits. We envision U-aggregation as a powerful tool for real-world implementation of machine learning models, particularly in the era of open science.

We adopt AMP for the recovery of $\bu$ and $\bv$ due to its strong theoretical foundation. However, the assumption of Gaussian-distributed noise may be restrictive in certain practical scenarios. As demonstrated in Figure S2 of the Supplementary Material, U-aggregation remains stable even when noise is generated from uniform distributions. Recent theoretical advancements suggest the universality of AMP theory beyond Gaussian noise \citep{wang2024universality}. In real-world applications, the AMP procedure could potentially be replaced by other iterative methods. Additionally, when sparsity is low, thresholding may no longer be necessary.

The proposed U-aggregation method is primarily motivated by the data model (\ref{datamodel}), which may not hold in certain applications. A key assumption underlying the method is the rank-one structure of the signal. This assumption can be assessed by examining the empirical singular values, either before or after variance stabilization, as suggested by \citet{landa2023dyson} and \citet{donoho2023screenot}. Our data application serves as a reality check, providing empirical support for the validity of this assumption in real-world problems. 

Moving forward, when integrating pre-trained models, external information regarding the reliability or similarity between the source training data and the target population may be available. Such external information can be incorporated into the procedure, for example, by imposing constraints on $\bu$ or providing insights into the variance structure. U-aggregation, along with other potential approaches for regular validation of model performance, coupled with feedback to model developers, may foster a positive learning cycle. This iterative process can enhance the quality of models and promote transparent, accurate performance evaluations, ultimately benefiting the data science community.

\section*{Data and Code Availability}
The code to implement the U-aggregation in the paper is available on GitHub, \url{https://github.com/biostat-duan-lab/Uaggregation} and data from All of Us research Hub are available for registered users from \url{https://www.researchallofus.org/data-tools/workbench/.} 

\section*{Acknowledgment}
This work was supported by National Institutes of Health (R01GM148494).

	\bibliographystyle{chicago}
	\bibliography{reference}

\end{document}